\documentclass[conference]{IEEEtran}
\usepackage{tabularx}
\usepackage{array, boldline, booktabs}
\usepackage{multirow}
\usepackage{amsmath,amssymb,amsfonts}
\usepackage{algorithmic}
\usepackage{graphicx}
\usepackage{textcomp}
\usepackage{slashbox}
\usepackage{caption}
\usepackage{xcolor}
\usepackage{hyperref, etoolbox}
\usepackage{float}
\usepackage{bm}
\usepackage{algorithm,algorithmic}
\usepackage{amsmath,amsfonts,amsthm}
\usepackage{mathrsfs}
\def\BibTeX{{\rm B\kern-.05em{\sc i\kern-.025em b}\kern-.08em
    T\kern-.1667em\lower.7ex\hbox{E}\kern-.125emX}}

\newcommand\Tstrut{\rule{0pt}{2.6ex}}         
\newcolumntype{P}[1]{>{\centering\arraybackslash}p{#1}}
\newcolumntype{M}[1]{>{\centering\arraybackslash}m{#1}}

\newcommand{\xx}{\bm{x}}
\newcommand{\yy}{\bm{y}}

\begin{document}

\title{Dual Head Adversarial Training}

\author{\IEEEauthorblockN{Yujing Jiang\textsuperscript{1}, Xingjun Ma\textsuperscript{2}, Sarah Monazam Erfani\textsuperscript{1}, James Bailey\textsuperscript{1}}
\IEEEauthorblockA{\textsuperscript{1}\textit{School of Computing and Information Systems, The University of Melbourne}\\
\{yujingj@student., sarah.erfani@, baileyj@\}unimelb.edu.au}

\IEEEauthorblockA{\textsuperscript{2}\textit{School of Information Technology, Deakin University}\\
daniel.ma@deakin.edu.au}
}

\maketitle

\begin{abstract}
Deep neural networks (DNNs) are known to be vulnerable to adversarial examples/attacks, raising concerns about their reliability in safety-critical applications. A number of defense methods have been proposed to train robust DNNs resistant to adversarial attacks, among which adversarial training has so far demonstrated the most promising results. However, recent studies have shown that there exists an inherent tradeoff between accuracy and robustness in adversarially-trained DNNs. In this paper, we propose a novel technique Dual Head Adversarial Training (DH-AT)  to further improve the robustness of existing adversarial training methods. Different from existing improved variants of adversarial training, DH-AT modifies both the architecture of the network and the training strategy to seek more robustness. Specifically, DH-AT first attaches a second network head (or branch) to one intermediate layer of the network, then uses a lightweight convolutional neural network (CNN) to aggregate the outputs of the two heads. The training strategy is also adapted to reflect the relative importance of the two heads. We empirically show, on multiple benchmark datasets, that DH-AT can bring notable robustness improvements to existing adversarial training methods. Compared with TRADES, one state-of-the-art adversarial training method, our DH-AT can improve the robustness by 3.4\% against PGD$^{40}$ and 2.3\% against AutoAttack, and also improve the clean accuracy by 1.8\%.
\end{abstract}

\section{Introduction}
Deep neural networks (DNNs) have been adopted to achieve state-of-the-art performance in a wide range of applications, such as computer vision \cite{he2016deep}, natural language processing \cite{devlin2018bert} and speech recognition \cite{amodei2016deep}. Despite the great success, DNNs have also been found to be extremely vulnerable to adversarial examples/attacks \cite{szegedy2013intriguing,goodfellow2014explaining}.
With imperceptible but carefully-crafted perturbations, natural (clean) examples can be converted into adversarial examples to fool state-of-the-art DNNs \cite{carlini2017towards,madry2017towards}. In recent research works, adversarial attacks have been demonstrated to be destructive to almost all kinds of DNNs including image models \cite{madry2017towards,carlini2017towards,wu2020skip}, video models \cite{jiang2019black}, graph models \cite{dai2018adversarial} and even language models like BERT \cite{sun2020adv}.
This has raised security concerns on the deployment of DNNs in safety-critical applications such as face recognition \cite{goswami2018unravelling}, autonomous driving \cite{eykholt2018robust,kong2020physgan,duan2020adversarial,tu2020physically,cao2019adversarial}, medical diagnosis \cite{finlayson2019adversarial,ma2021understanding} and many others.

A number of methods have been proposed to defend DNNs against adversarial attacks adversarially robust DNNs \cite{guo2017countering,ma2018characterizing,dhillon2018stochastic,xie2018mitigating,madry2017towards}, among which adversarial training (AT) has demonstrated the most promising results \cite{athalye2018obfuscated,croce2020reliable,jiang2020imbalanced}.
Adversarial Training (AT)~\cite{goodfellow2014explaining} involves adversarial samples in each training step to enhance the model's robustness, which can be formulated as a min-max optimization problem \cite{madry2017towards,wang2019convergence}. Most existing adversarial training methods adopt WideResNets (WRNs) \cite{zagoruyko2016wide} to demonstrate the best robustness results. 
By upscaling ResNets (RNs) at the width dimension, WRNs introduce more capacity into RNs in an efficient manner, which has been found to be crucial for adversarial robustness \cite{zhang2019theoretically,wang2019improving}. It has been observed that using WRNs can bring consistent robustness improvement over RNs \cite{zhang2019theoretically,wang2019improving,wu2020does}. 
So far, the two most commonly used WRNs are WRN-34-10 \cite{madry2017towards,zhang2019theoretically,huang2020self,wu2020adversarial} and WRN-34-20 \cite{rice2020overfitting,pang2020bag,pang2020boosting,cui2020learnable} ($\sim$4$\times$ more parameters than WRN-34-10).
In this paper, we aim to explore a more efficient and effective way to improve adversarial robustness with adversarial training that does not significantly increase the model's capacity.

Following the standard adversarial training (SAT) proposed by Madry et al. \cite{madry2017towards}, many adversarial training techniques introduce new loss functions or training strategies with additional tunable hyperparameters to improve robustness under different settings~\cite{pang2020bag}. For example, TRADES \cite{zhang2019theoretically} adopts a hybrid of the cross-entropy (CE) and the Kullback–Leibler (KL) divergence loss with a balancing hyperparameter ($\lambda$) to explore different trade-offs between the clean accuracy and adversarial robustness. This has been found to be an important generalization of SAT with substantial ($> 5\%$) robustness improvement \cite{zhang2019theoretically}.
A closer look at TRADES under different $\lambda$s, we find  that the adversarial noise patterns generated from these models are quite different on test examples, and the transferability of these patterns to other models is very limited. Moreover, the attack will be significantly weakened if we generate adversarial examples using the averaged perturbation over two TRADES models trained with different $\lambda$s. These results indicate that different $\lambda$s produce models that are robust in distinctive ways.
This motivates us to discover a novel technique that can exploit different training parameters in one single model via separate output heads, and effectively aggregate those heads to yield a more robust model.

In this paper, we propose \emph{Dual Head Adversarial Training} (DH-AT), an improved variant of AT that attaches a second head to one intermediate layer of the network. In WRNs, the second head can be symmetrically attached to the end of the first residual block (illustrated in Fig. \ref{fig:wrn}). When TRADES~\cite{zhang2019theoretically} training method is considered, the two heads can be trained either simultaneously or independently with different $\lambda$s. The main (existing) head can also be directly loaded from a pre-trained model without any modifications, in which case only one head requires training. After training the two heads, a lightweight convolutional neural network (CNN) can then be adversarially trained to combine the two heads, which takes fewer than 20 epochs. In real-world scenarios, the second head and the lightweight CNN together form a strengthening mechanism to improve the adversarial robustness of any existing models. Note that the second head can also be switched off when robustness is no longer the primary concern.

In summary, our main contributions are:

\begin{itemize}
    \item We propose a novel Dual Head Adversarial Training (DH-AT) method to improve the adversarial robustness of any existing models by attaching a second output head to the network. DH-AT can be easily incorporated into existing adversarial training methods with minimal modifications.

  \item Our DH-AT provides a novel defense strategy with one head is responsible for clean accuracy and the other head for adversarial robustness. With our DH-AT, achieving both clean accuracy and robustness at the same time is possible, as evidenced by our experimental results.

  \item Following DH-AT, we demonstrate a novel alternative to early-stopping by using the best epoch's weights as the main head and a ``robust overfitted"~\cite{rice2020overfitting} as the second head. On CIFAR-10, this leads to up to 2.01\% robustness improvement against PGD$^{40}$ and 1.19\% against AutoAttack~\cite{croce2020reliable} on CIFAR-10, when compared with TRADES.
  

 
\end{itemize}

\section{Related Work}
In this section, we briefly review existing works in both adversarial attack and defense. We focus on those methods that were developed in the white-box setting where model parameters are known to either the attacker or defender. And we will focus on those developed with image classification models, the major field of adversarial research.

\subsection{Adversarial Attack}
The vulnerability of DNNs to small adversarial perturbations was initially discovered by Szegedy et al. \cite{szegedy2013intriguing}, where adversarial examples were crafted to fool state-of-the-art classification DNNs.
By far, a significant amount of research has been conducted to either design more powerful adversarial attacking methods to examine the robustness of different DNNs.
The most classic and efficient attacking method is the Fast Gradient Sign Method (FGSM)~\cite{goodfellow2014explaining}, which only takes one single step of gradient ascent to maximize the model's classification error. An iterative version of FGSM was then proposed to enhance the attack strength in physical-world scenarios.  This attack is known as Basic Iterative Method (BIM) \cite{kurakin2016adversarial}. Proposed by Madry et al. \cite{madry2017towards}, the Projected Gradient Descent (PGD) has been recognized as one of the strongest first-order adversarial attacks. PGD iteratively perturbs the input sample $\xx$ with a smaller step size and clips the perturbation back to an $\epsilon$-ball around $\xx$ if it goes beyond. The $\epsilon$ perturbation constraint is  defined by the $l_{\infty}$ norm.
Other well-known and effective adversarial attacks include DeepFool~\cite{moosavi2016deepfool}, Carlini and Wagner (CW) attacks~\cite{carlini2017towards}, Jacobian-based Saliency Map Approach (JSMA)~\cite{papernot2016limitations}, Momentum Iterative Attack~\cite{dong2018boosting}, Distributionally Adversarial Attack~\cite{zheng2019distributionally} and Margin Decomposition (MD) attacks \cite{jiang2020imbalanced}.
Recently, Croce et al. \cite{croce2020reliable} proposed the AutoAttack (AA) which is a parameter-free ensemble of four different adversarial attack methods.  AA has been shown to be the most reliable attack for robustness evaluation.

\subsection{Adversarial Defense}
A wide range of defense methods have been proposed to improve DNN adversarial robustness, such as defensive distillation \cite{papernot2016distillation}, adversarial detection \cite{xu2017feature,ma2018characterizing}, input denoising \cite{guo2018countering,niu2020limitations}, gradient regularization \cite{ross2018improving,jakubovitz2018improving,finlay2021scaleable}, model compression \cite{ye2019adversarial,guo2018sparse} and adversarial training (AT) \cite{goodfellow2014explaining,madry2017towards}. While some these defense methods are still vulnerable to adaptive attacks \cite{athalye2018obfuscated,jiang2020imbalanced}, AT methods have been found to be the most reliable defense. 

AT gains robustness by training the model on adversarial (instead of clean) examples at each training iteration \cite{madry2017towards,wang2019convergence}. 
Training with the PGD adversarial examples is known as the Standard Adversarial Training (SAT) \cite{madry2017towards}. SAT is the defense method that for the first time can bring considerable robustness into DNNs.
A number of variants of SAT have recently been proposed to further improved the robustness of SAT.
Ensemble adversarial training~\cite{tramer2017ensemble} loosens the model’s decision boundary and augments clean training examples with perturbations transferred from diverse models.
TRADES~\cite{zhang2019theoretically} exploits the KL distance between the model's outputs on clean versus adversarial examples to generate stronger (than PGD) attacks, thus can train more robust models. The objective of TRADES has two loss terms with one is the commonly used CE loss defined on clean examples and the other one is the KL divergence between the model's output on clean versus adversarial examples. It uses a regularization hyperparameter $\lambda$ to exert different trade-offs between the clean accuracy and adversarial robustness. According to recent evaluations, TRADES improves the robustness of SAT by $\sim5\%$ on CIFAR-10, which is a very significant improvement. Recently, Wang et al. \cite{wang2019improving} proposed the Misclassification Aware adveRsarial Training (MART), which is a variant of TRADES that improves robustness by differentiating misclassified from correctly-classified examples. 

The recent trend of adversarial training also involves exploiting deeper or wider network architectures for better adversarial robustness, such as  WRN-34-10 \cite{madry2017towards,zhang2019theoretically,huang2020self,wu2020adversarial,bai2021improving}, WRN-34-20 \cite{rice2020overfitting,pang2020bag,pang2020boosting,cui2020learnable} or even WRN-106-8 \cite{alayrac2019labels}. Compared to WRN-34-10, WRN-34-20 and WRN-106-8 consist of $\sim$4$\times$ and $\sim$2.3$\times$ parameters, respectively. This tends to incur a significant amount of more training time. In this work, we explore smarter ways to improve robustness by attacking separate heads/branches to existing networks rather than simply scaling up the entire architecture like WRN-34-20 or WRN-106-8. 
Based on our analysis, the proposed DH-AT only introduces $\sim$0.95$\times$ more parameters to WRN-34-10 at the cost of a linearly increased training time by $\sim$1.2$\times$. 
This has led to much more robustness improvement than scaling WRN-34-10 up to WRN-34-20 or WRN-106-8. Moreover, it has been shown that clean accuracy may be inherently at odds with adversarial robustness \cite{tsipras2018robustness}.

In this work, we propose the use of separate heads in DNNs to achieve two purposes: 1) combining different levels of robustness into one single model via different heads; and 2) better trade-off between clean accuracy and adversarial robustness with each head is responsible for only one property (\textit{e.g.}, accuracy or robustness). This admits more flexible defense strategies in real-world scenarios, for example, the robustness head can be switched off when robustness is no longer the primary concern, or secure a higher level of robustness using the more robust head or both. The most relevant work to ours is the recent use of adversarial examples along with a separate auxiliary batch norm to improve image recognition \cite{xie2020adversarial}. However, we focus more on the output of the network and its adversarial robustness, and more importantly, in the adversarial not the clean training setting. 
From the ensemble perspective, WRN-34-20 is a costly ensemble of two WRN-34-10 models, however, our dual head strategy provides a smarter way for ensembling. In other words, improved robustness can be achieved by simply ensembling multiple heads (rather than the entire network) into one single network.

\section{Dual Head DNNs and Dual Head Adversarial Training}
\label{sec:DH-AT}
In this section, we first explain our intuition for designing dual head models, then describe how to design a dual head model for RNs or WRNs. Finally, we introduce the proposed dual head adversarial training strategy.

\subsection{Intuition for Using Two Heads}

\begin{figure}[!h]
\centering
\begin{tabular}{M{68pt}M{6pt}M{68pt}M{68pt}}
 \textbf{Original Image}  &   & \textbf{76$^{th}$ Epoch} (Example 1) & \textbf{91$^{st}$ Epoch} (Example 2)\\ \Tstrut
 \includegraphics[width=68pt]{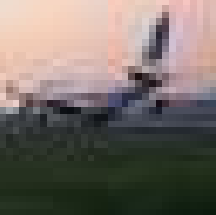}  & \rotatebox[origin=lb]{90}{Perturbed} &  \includegraphics[width=68pt]{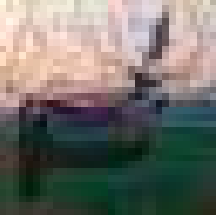}    &    \includegraphics[width=68pt]{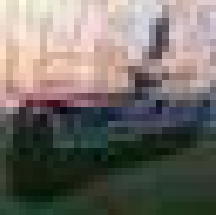}  \\
 & \rotatebox[origin=lb]{90}{Adv. Noise} & \includegraphics[width=68pt]{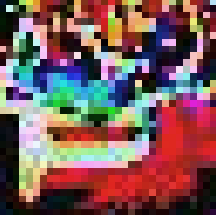}              &        \includegraphics[width=68pt]{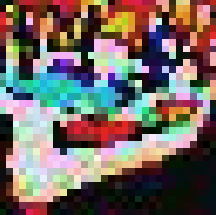}
\end{tabular}
\begin{tabular}{c|cc} \Tstrut
 \backslashbox{Input}{Checkpoint}    & 76$^{th}$ Epoch  & 91$^{st}$ Epoch  \\ \hline
Adv. example 1 & Airplane  &  Bird     \\
Adv. example 2 & Ship  &  Airplane
\end{tabular}
\caption{Adversarial examples (top row, right two columns) and noises (middle row, right two columns) generated by PGD$^{100}$ ($\epsilon=8/255$) for the same clean image (top row, left column) on different epochs of checkpoint. The model adopts WRN-34-10 trained with TRADES ($1/\lambda=6$) on CIFAR-10 dataset (with image size $32\times32$), and the checkpoints are captured from the 76$^{th}$ and 91$^{st}$ epochs. The table reports the model's predictions on the two adversarial examples (true label is `airplane'). The first adversarial example is generated from the 76$^{th}$ epoch, while the second example is generated from the same model at the 91$^{st}$ epoch. For better illustration, the adversarial noises are enhanced by 20$\times$.}
\label{fig:es_test}
\end{figure}

A recent study~\cite{rice2020overfitting} shows that some AT methods heavily rely on early-stopping to achieve top-ranked robustness. Rice et al. \cite{rice2020overfitting} demonstrated that a similar performance gain can be achieved by a piece-wise learning rate scheduler along with smart early-stopping, which both can prevent ``robust overfitting". In our investigation, TRADES~\cite{zhang2019theoretically} could achieve 55.88\% robustness against PGD$^{40}$ ($\epsilon=8/255$) on CIFAR-10 dataset~\cite{rice2020overfitting}. However, with only 15 more epochs of training after the best checkpoint, the robustness under the same attack drops by 2\%+. This phenomenon leads us to inspect the differences in adversarial noises generated by PGD on models obtained at the best versus the last checkpoints. 
As shown in Fig. \ref{fig:es_test}, the adversarial noises (right two columns) generated for the same clean image (left column) are notably different for models obtained at the 76$^{th}$ versus the 91$^{st}$ training epochs. This phenomenon is consistent across different CIFAR-10 test images. We also find that the adversarial noise generated on one model does not necessarily cause the same error on the other (see the prediction table in Fig. \ref{fig:es_test}). 
This indicates that the two models have different levels of robustness and they are robust in distinctive ways. This phenomenon can also be expected to exist between models trained using different parameters. This motivates us to explore ways to combine different levels of robustness in one single model. Since robustness is more reliable at a later training stage when the shallow layers of the network are less likely to be significantly updated, we propose to use additional output branches (\textit{i.e.}, heads) to include more
levels of robustness into a single network, \textit{i.e.}, a dual head architecture.

\subsection{Dual Head DNNs}
\begin{figure}[!h]
\centering
\includegraphics[width=\linewidth]{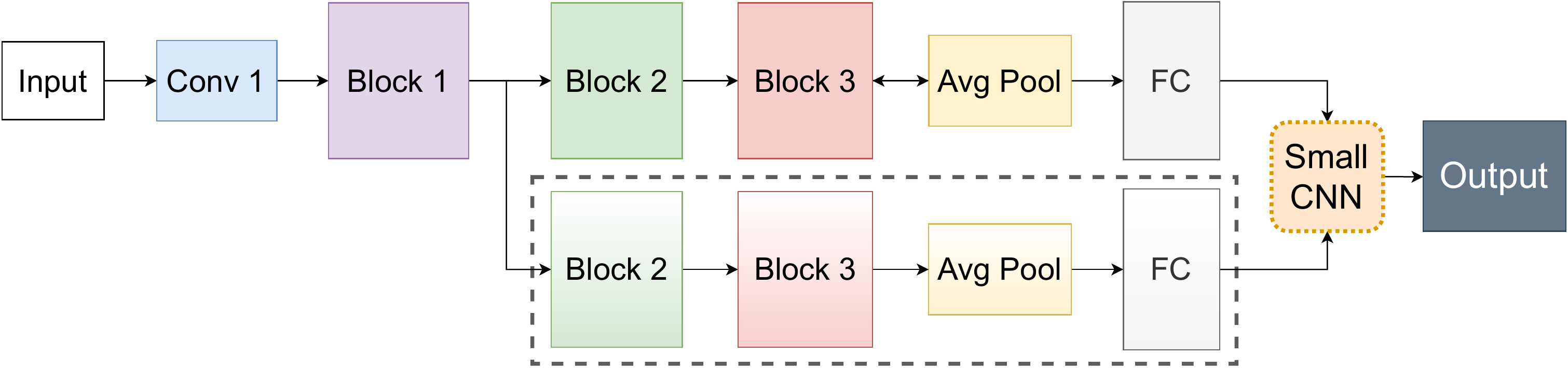}
\caption{The symmetric dual head architecture for WRN-34-10}
\label{fig:wrn}
\end{figure}

We propose two types of duel head architectures: symmetric and asymmetric.
The symmetric dual head architecture for WRN-34-10 is illustrated in Fig. \ref{fig:wrn}. In this architecture, each head traverses the complete structure of a WRN-34-10 network, while sharing the first convolutional layer (Conv 1) and the first residual block (Block 1). In other words, the original model branches into two heads after the first residual block (Block 1), where we denote Block 1 as the ``attaching point" of the model. To maximally employ convolutional layers that recognize higher-level features, the attaching point is preferably selected after the first group of repeated convolutional layers with identical filter arrangements. Particularly, for WRNs, we choose the output of Block 1 as the attaching point, regardless of the network's depth and width.

\begin{figure}[!h]
\centering
\includegraphics[width=\linewidth]{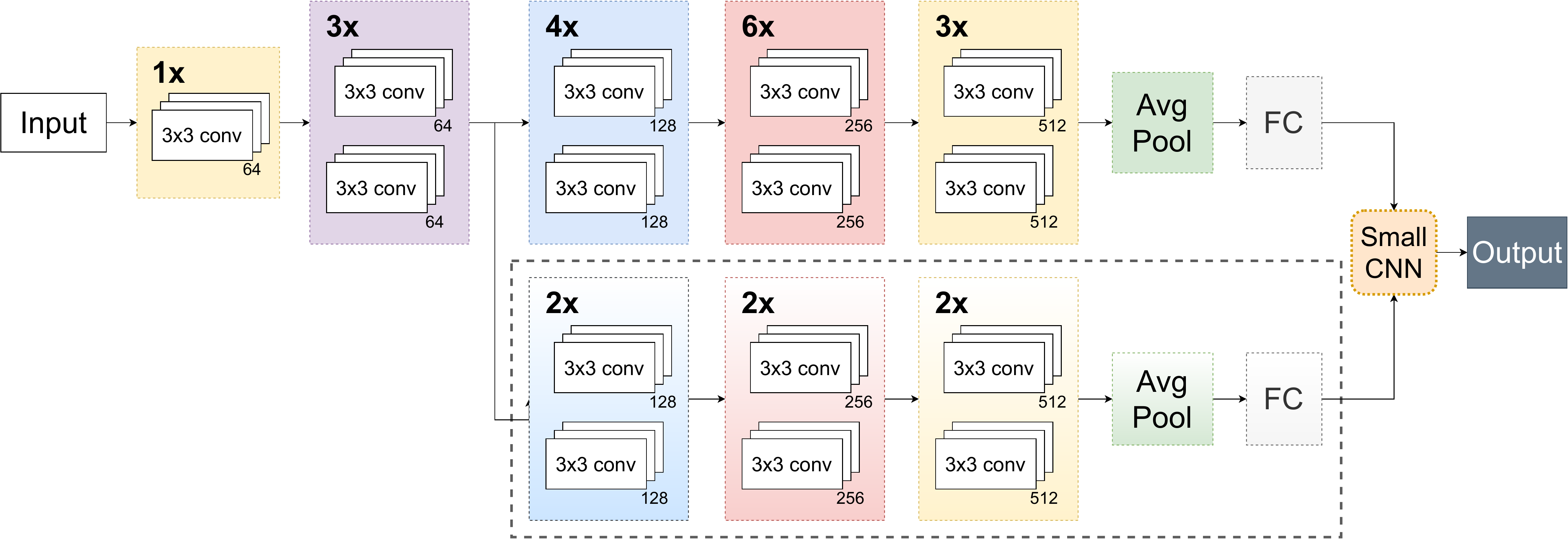}
\caption{The asymmetric dual head architecture for WRN-34-10}
\label{fig:rn}
\end{figure}

Asymmetric dual head architectures employ different depths of ResNets for the two heads. As illustrated in Fig. \ref{fig:rn}, the original ResNet-34 network serves as the main head and a revised ResNet-18  network is attached as the second head. Here, the attaching point is selected to be the output of the first residual group (\textit{i.e.}, the 4$^{th}$ convolutional block of the entire network). Note that, this asymmetric dual head version of WRN-34-10 employs all 3 residual blocks in the first residual group for both heads, while the standard ResNet-18 only employs 2 of them. After the attaching point, the second head (enclosed in the dashed box) has approximately 50\% of the parameter size of the main head.

\begin{figure}[h]
\centering
\includegraphics[width=\linewidth]{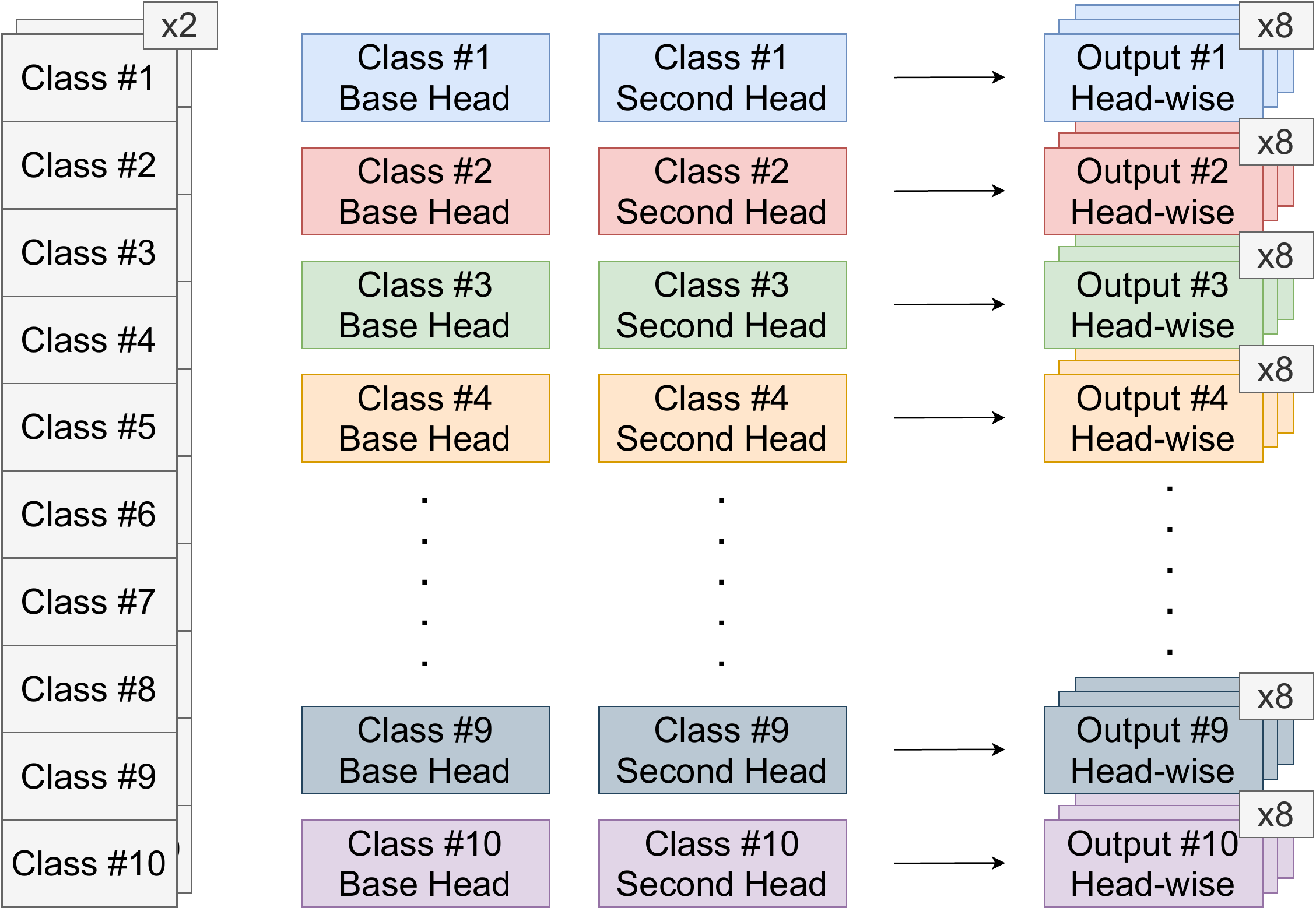}
\caption{Head-wise logits convolution}
\label{fig:conv1}
\centering

\end{figure}

\begin{figure}[h]
\centering
\includegraphics[width=\linewidth]{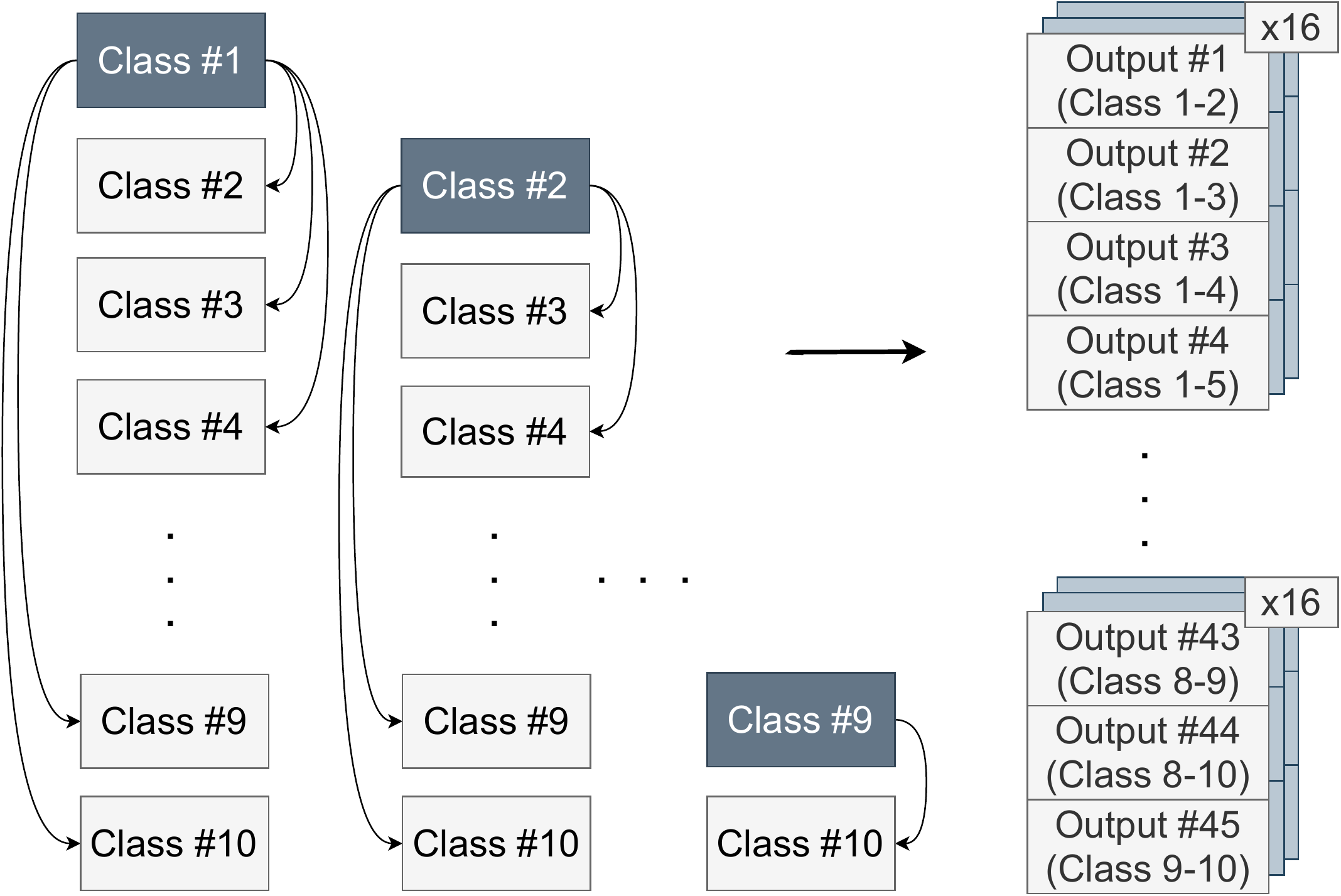}
\caption{Class-wise feature-map convolution}
\label{fig:conv2}
\centering

\end{figure}

Different from conventional multi-head networks like the Siamese network \cite{bromley1993signature}, here we further combine the two heads into one single specifically designed output subnetwork.
We design a lightweight CNN to aggregate the output from each head. The CNN subnetwork comprises two types of conventional filters: head-wise logits convolution (shown in Fig. \ref{fig:conv1}) and class-wise logits convolution (shown in Fig. \ref{fig:conv2}), where the logits refer to the output of the fully connected (FC) layers (no softmax). The logits output of each head (originally being two 1$\times$10 arrays) will be merged into a 1$\times$10$\times$2 tensor before feeding into the lightweight CNN (illustrated in the left part of Fig. \ref{fig:conv1}). For both types of convolution, they use the same 1$\times$2 convolutional kernels, except traversing through different dimensions.

\begin{figure}[H]
\centering
\includegraphics[width=\linewidth]{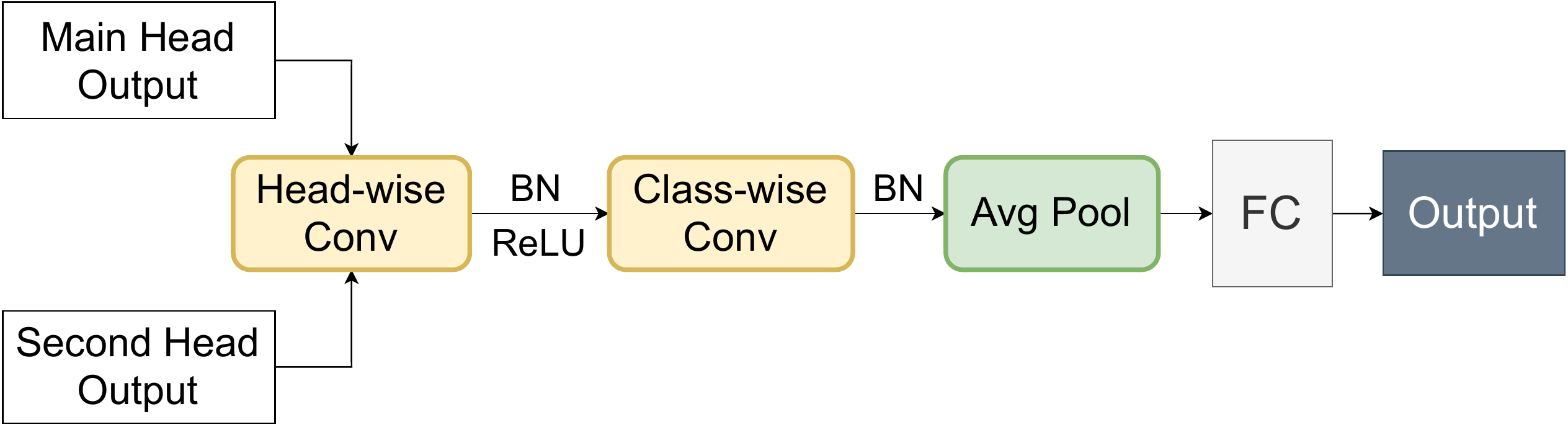}
\caption{The architecture of the lightweight CNN}
\label{fig:smallcnn}
\centering
\end{figure}

The head-wise logits convolution applies 8 convolutional kernels with the size of 1$\times$1$\times$2 to traverse across different classes. For each convolutional operation, it involves logits from each head regarding the same class (examples can be found in Fig. \ref{fig:conv1}). Accordingly, a feature map of size 8$\times$10$\times$1 will be generated for each input. This head-wise logits convolution is intuitively designed to associate the logits of each class from the two heads. Then, batch normalization and ReLU activation are performed on the feature maps.

For the class-wise feature-map convolution, it utilizes 16 convolutional kernels with size 1$\times$2$\times$1 to traverse the feature maps previously generated by the head-wise logits convolution. The class-wise feature-map convolution is different from ordinary convolutional kernels. Fig. \ref{fig:conv2} shows an example with 10 classes. It will be applied to logits in all 45 combinations of classes in the first dimension of the input feature map (\textit{i.e.}, the dimension of the kernels in the previous convolution, namely the one with the size of 8). Accordingly, a new feature map of size 16$\times$8$\times$45$\times$1 will be generated in this layer. The class-wise convolution enables the network to learn the class-wise correlations between clean and adversarial examples. For instance, a clean example with ground truth Class \#3 may be predicted to be Class \#7 after adversarial perturbation. Such class cross information is important for adversarial robustness, yet has been overlooked in existing adversarial training methods.

Then, we use an average pooling layer with stride 2 to reduce the feature maps' dimensionality along the second dimension, the same dimension where class-wise feature-map convolution is applied (the one with size 16). After this pooling layer, the dimension of the feature maps will be reduced to 8$\times$8$\times$45. Finally, the feature maps will be flattened to size 2880$\times$1 before passing into a fully connected (FC) layer (with softmax activation) to output the final predictions.

\subsection{Dual Head Adversarial Training}
\label{sec:training}

The training procedure of DH-AT associates the three components in a particular order: the main head, then the second head, and finally the lightweight CNN. Given a specific adversarial training method and a dual head DNN, DH-AT first adversarially trained the main head from scratch. Note that, any pre-trained model can be used as the main head.
Next, we attach the second head to the main head at the specified attaching point and freeze the parameters of every component of the main head before the attaching point. 
We then train the second head using the same adversarial training method but with different hyperparameter settings, for example, by altering the attack intensity $\epsilon$ or changing loss hyperparameters (\textit{e.g.}, in TRADES or MART). 
The last step of training is to train the lightweight CNN as follows:
\begin{equation}
\label{equ:trades_loss}
\min_{f} \mathbb{E} \{CE(f(\xx),\yy)+\max_{\xx'\in\mathbb{E}(\xx,\epsilon)} KL(f(\xx),f(\xx'))/\lambda\},
\end{equation}
where $\xx'$ is the adversarial example generated using PGD for clean example $\xx$, and $\lambda$ with $1/\lambda=2$ is balancing parameter on the adversarial output of the lightweight CNN. 
Note that the above loss is also the loss function used in TRADES~\cite{zhang2019theoretically}. 
In this training stage, each training clean sample is paired with its PGD adversarial example which together is passed into the network in batches for model training. 
On CIFAR-10, when WRN-34-10 is used as the main head, the lightweight CNN can converge within 20 epochs of adversarial training.

Our dual head design and DH-AT training strategy provide a flexible way to keep two different levels of robustness in one single model. For example, keeping the best and the last checkpoints of the same model, or keeping models trained using different methods or with different hyper-parameters. Taking the best-last checkpoint case, for example, one can first train the main head using an existing adversarial training method, and at the same time, monitoring the validation robustness on a small validation set. When the validation robustness starts to drop, stop training the network and leave it as the main head. We can then copy all the weights to the second head, freeze the parameters of the main head before the attaching point and train the second head for a certain number of epochs until the training loss converges. We can then free both heads to train the lightweight CNN, again using the same adversarial training method. Note that, due to the special design of the lightweight merging CNN, the second head can be easily switched on and off to meet different application scenarios, which is also the case for the main head.



\section{Experiments: Symmetric DH-AT with WRN-34-10}

\begin{table*}[!t]
\caption{Robustness results of different defense methods under various parameter settings, on CIFAR-10 data. For all DH-AT models (WRN-34-10s), the left cell shows the results for the individual head (upper: main head; lower: second head), while the right cell shows the result for the final model. Robustness is evaluated on CIFAR-10 test set against $\epsilon=8/255$ attacks.}
\centering
\setlength{\tabcolsep}{0.8em}
\begin{tabular}{p{115pt}cccccccccc}
\toprule
   & \multicolumn{2}{c}{\textbf{Natural}} & \multicolumn{2}{c}{\textbf{PGD$^{40}$}} & \multicolumn{2}{c}{\textbf{PGD$^{100}$}} & \multicolumn{2}{c}{\textbf{APGD-DLR-T}} & \multicolumn{2}{c}{\textbf{AutoAttack}} \\ \midrule
TRADES~\cite{zhang2019theoretically}               & \multicolumn{2}{c}{84.97\%}            & \multicolumn{2}{c}{55.88\%}          & \multicolumn{2}{c}{55.64\%}           & \multicolumn{2}{c}{53.10\%}                & \multicolumn{2}{c}{53.08\%}               \\ \midrule
TRADES $1/\lambda=6.0$ (76$^{th}$ Epoch)    & 84.97\%    & \multirow{2}{*}{86.78\%}    & 55.88\%   & \multirow{2}{*}{\textbf{59.31\%}}   & 55.64\%    & \multirow{2}{*}{\textbf{58.92\%}}   & 53.10\%       & \multirow{2}{*}{\textbf{56.84\%}}     & 53.08\%      & \multirow{2}{*}{\textbf{55.38\%}}     \\
TRADES $1/\lambda=3.0$ (76$^{th}$ Epoch)    & 86.45\%    &                           & 53.17\%    &                          & 52.65\%    &                          & 50.36\%      &                            & 50.31\%       &                            \\ \midrule
TRADES $1/\lambda=3.0$ (78$^{th}$ Epoch)     & 86.97\%    & \multirow{2}{*}{\textbf{87.97\%}}    & 53.30\%    & \multirow{2}{*}{57.34\%}   & 52.92\%    & \multirow{2}{*}{57.01\%}   & 50.93\%      & \multirow{2}{*}{54.25\%}     & 50.80\%       & \multirow{2}{*}{53.41\%}     \\
TRADES $1/\lambda=2.0$ (78$^{th}$ Epoch)    & 87.73\%    &                           & 51.73\%   &                          & 51.37\%    &                          & 49.78\%      &                            & 49.64\%      &                            \\ \midrule
TRADES $1/\lambda=6.0$ (76$^{th}$ Epoch)     & 84.97\%    & \multirow{2}{*}{85.93\%}    & 55.88\%   & \multirow{2}{*}{57.89\%}   & 55.48\%    & \multirow{2}{*}{57.58\%}   & 53.10\%      & \multirow{2}{*}{54.63\%}     & 53.08\%      & \multirow{2}{*}{54.27\%}     \\
TRADES $1/\lambda=6.0$ (91$^{st}$ Epoch)     & 85.87\%    &                           & 53.52\%    &                          & 53.06\%    &                          & 52.75\%     &                            & 52.53\%       &                            \\ \midrule
TRADES $1/\lambda=3.0$ (78$^{th}$ Epoch)     & 86.97\%     & \multirow{2}{*}{87.63\%}    & 53.30\%   & \multirow{2}{*}{56.08\%}   & 52.92\%    & \multirow{2}{*}{55.62\%}   & 50.93\%      & \multirow{2}{*}{53.03\%}     & 50.80\%      & \multirow{2}{*}{52.84\%}     \\
TRADES $1/\lambda=3.0$ (93$^{th}$ Epoch)     & 87.50\%    &                           & 51.55\%    &                          & 50.91\%    &                          & 50.19\%      &                            & 50.07\%       &                            \\ \midrule \midrule

SAT~\cite{madry2017towards}                & \multicolumn{2}{c}{87.32\%}            & \multicolumn{2}{c}{46.75\%}          & \multicolumn{2}{c}{45.26\%}           & \multicolumn{2}{c}{44.78\%}               & \multicolumn{2}{c}{44.23\%}               \\ \midrule
SAT (160$^{th}$ Epoch)         & 87.32\%   & \multirow{2}{*}{87.84\%}    & 46.75\%   & \multirow{2}{*}{\textbf{47.81\%}}   & 45.26\%    & \multirow{2}{*}{\textbf{47.24\%}}   & 44.78\%      & \multirow{2}{*}{\textbf{45.48\%}}     & 44.23\%      & \multirow{2}{*}{\textbf{45.06\%}}     \\
SAT (175$^{th}$ Epoch)         & \textbf{87.98\%}   &                           & 45.82\%   &                          & 45.01\%    &                          & 44.52\%      &                            & 43.96\%      &                            \\ \midrule \midrule

MART~\cite{wang2019improving}                 & \multicolumn{2}{c}{84.15\%}            & \multicolumn{2}{c}{58.25\%}          & \multicolumn{2}{c}{57.78\%}           & \multicolumn{2}{c}{55.24\%}               & \multicolumn{2}{c}{54.68\%}               \\  \midrule

MART $\lambda=6.0$ (84$^{th}$ Epoch)       & 84.15\%    & \multirow{2}{*}{\textbf{85.51\%}}    & 58.25\%   & \multirow{2}{*}{\textbf{59.94\%}}   & 57.78\%    & \multirow{2}{*}{\textbf{59.36\%}}   & 55.04\%      & \multirow{2}{*}{\textbf{56.23\%}}     & 54.68\%      & \multirow{2}{*}{\textbf{55.75\%}}     \\
MART $\lambda=3.0$ (84$^{th}$ Epoch)        & 85.48\%    &                           & 56.83\%   &                          & 56.17\%    &                          & 53.97\%      &                            & 53.11\%      &        \\  \midrule
MART $\lambda=6.0$ (84$^{th}$ Epoch)      &  84.15\%   & \multirow{2}{*}{84.52\%}    &  58.25\%  & \multirow{2}{*}{59.06\%}   &  57.78\%  & \multirow{2}{*}{58.57\%}   &    55.24\%   & \multirow{2}{*}{55.36\%}     &   54.68\%    & \multirow{2}{*}{55.23\%}     \\
MART $\lambda=6.0$ (99$^{th}$ Epoch)      & 84.29\%    &                           & 57.23\%   &                          & 56.84\%    &                          & 54.60\%  &                            & 54.02\%      &       \\                   
\bottomrule
\end{tabular}
\label{tbl:cifar10_results}
\end{table*}

We first test our symmetric DH-AT in which the two heads of the network are identical. All baseline models (SAT, TRADES, and MART) are trained following their settings specified in the original papers, except for utilizing 2 GPUs in parallel. During training, we apply PGD$^{10}$ with $\epsilon=8/255$ and step size $2/255$ to generate training adversarial examples. We evaluate all baseline and our models using 1) untargeted PGD attack, 2) targeted Auto-PGD (APGD) attack~\cite{croce2020reliable}, and AutoAttack~\cite{croce2020reliable} which is an ensemble of 4 different attacks including the APGD. 
We follow existing works to use the same $\epsilon=8/255$ for testing for all attacks. The step sizes for PGD attacks are set to $2.5\cdot\epsilon/num\_steps$ when different perturbation steps are considered. The complete robustness results are summarized in Table \ref{tbl:cifar10_results}. Next, we will detail these results according to the adversarial training method used.

\subsection{Detailed Experimental Settings}

\subsubsection{DH-AT with TRADES}

To combine our DH-AT with TRADES, we investigate two different settings: 1) training the two heads with different $\lambda$ hyperparameters, and 2) training the two heads by different numbers of epochs. Same as the original TRADES, each head in our DH-AT model is trained for 100 epochs using Stochastic Gradient Descent (SGD) with initial learning rate 0.1, momentum 0.9, and weight decay $2e^{-4}$. The learning rate is decayed by a factor of $1/10$ at the 75$^{th}$, 90$^{th}$, and 100$^{th}$ epochs.

\subsubsection{Heads trained by TRADES with different $\lambda$s}
In this setting, the two heads are trained using TRADES with different $\lambda$ hyperparameters. We denote the $\lambda$ used by the main and second head as $\lambda_1$ and $\lambda_2$, and test two sets of combinations: 1) $1/\lambda_1=6$ and $1/\lambda_2=3$, and 2) $1/\lambda_1=3$ and $1/\lambda_2=2$.
The training of the two heads follows the procedure described in Section \ref{sec:training}.
With a small validation set (1000 CIFAR-10 test images), we find that the best checkpoint of TRADES is the 76$^{th}$ epoch.
We then freeze the main head to the 76$^{th}$ epoch, and continue to train the second head using $1/\lambda_2=3$ for another 76 epochs. After this, we freeze both heads, attach the lightweight CNN and train the CNN for 15 epochs using TRADES loss with $1/\lambda=2$ and a learning rate of 0.02. 
Note that, for the second experiment with $1/\lambda_1=3$ and $1/\lambda_2=2$, the best checkpoint is found to be the 78$^{th}$ epoch. Apart from these changes, the remaining procedures are identical to the first experiment. We take the best checkpoints of the standalone TRADES as our baselines.
The results are reported in the top 2 - 5 rows in Table \ref{tbl:cifar10_results} (the first row is the standard TRADES result).

\subsubsection{Heads trained by TRADES for different epochs}

In this setting, we use the best checkpoint of TRADES as the main head, and a ``robust overfitted"~\cite{rice2020overfitting} subnetwork as the second head. The two heads are trained for different numbers of epochs, but using the same TRADES training technique with the same $\lambda$. Here, we test two different $\lambda$s: $1/\lambda=6$ and $1/\lambda=3$. The best checkpoints under the two $\lambda$s are the 76$^{th}$ and  78$^{th}$ epoch respectively.
The results are reported in the top 6 - 9 rows in Table \ref{tbl:cifar10_results}.

\subsubsection{DH-AT with SAT and MART}
We also experiment on DH-AT with SAT~\cite{madry2017towards} and MART~\cite{wang2019improving} to demonstrate the compatibility of DH-AT to different adversarial training methods.
Since SAT~\cite{madry2017towards} loss function does not have tunable parameters, we only apply DH-AT with different numbers of epochs for the two heads. 
For the original SAT, we train the networks for 200 epochs using SGD with an initial learning rate 0.1, momentum 0.9, and weight decay $2e^{-4}$. The learning rate is divided by 10 at the 100$^{th}$ and 150$^{th}$ epochs.
The best checkpoint at the 160$^{th}$ epoch is selected as the main head. The second head is trained for another 15 epochs.
The results are shown in rows 10 - 12 in Table \ref{tbl:cifar10_results}.

Improved from TRADES, MART~\cite{wang2019improving} also has a tuneable hyperparameter $\lambda$ in its loss function. 
Following the above DH-AT experiment with TRADES, here we train the main head using $1/\lambda_1=6$ for 85 epochs, then train the second head with $1/\lambda_2=3$ for the same number of epochs. 
We also evaluate DH-AT on MART ($1/\lambda=6$) with different numbers of epochs for the two heads, where the second head is trained for another 15 epochs after copied over from the main head, the best checkpoint obtained at the 85$^{th}$ epoch).
The results are reported in the bottom 5 rows in Table \ref{tbl:cifar10_results}.

\subsection{ Results Analysis}

Compared with standard TRADES ($1/\lambda=6$), using DH-AT with the two heads having different hyperparameters $1/\lambda_1=6$ and $1/\lambda_2=3$ demonstrates a considerable robustness improvement by 3.43\% against PGD$^{40}$ and 2.30\% against AutoAttack~\cite{croce2020reliable} (top 5 rows in Table \ref{tbl:cifar10_results}). Moreover, the clean accuracy is also improved by more than 2\%, which indicates that clean accuracy and robustness can be improved simultaneously. 
Note that, the training time of this type of DH-AT is approximately $2.2\times$ compared to standard TRADES. 
According to the results in rows 6 - 9 in Table \ref{tbl:cifar10_results}, applying our DH-AT with TRADES can effectively exploit the best and the last checkpoints, leading to 2.01\% robustness boost against PGD$^{40}$ and 1.19\% against AutoAttack. 
This verifies the effectiveness of our proposed DH-AT for overcoming the overfitting issue in adversarial training. 
Note that, in this setting, the training time of DH-AT is approximately $\sim$1.4$\times$ compared to standard TRADES. 
These two sets of experiments of DH-AT with TRADES demonstrate the flexibility of our DH-AT in different settings. It can be applied to incorporate two different levels of robustness or checkpoints into one single model. 

Results in the bottom 8 rows in Table \ref{tbl:cifar10_results} demonstrate the good compatibility of our DH-AT strategy with different adversarial training methods.
When utilizing MART with either $\lambda=6$ or $\lambda=3$ for both heads, our DH-AT method is able to deliver the highest adversarial robustness of 59.94\% (improved by 1.69\% compared to standard MART $\lambda=6$) against PGD$^{40}$ and 55.75\% (improved by 1.07\% compared to standard MART) against AutoAttack.
Meanwhile, the clean accuracy is also improved compared to standalone SAT/MART, though only slightly.
Since SAT uses the cross-entropy (CE) loss and the learning rate is decayed to 0.001 before the best checkpoint, its performance is relatively stable during additional training. However, applying our proposed DH-AT strategy can still improve its robustness by 1.06\% against PGD$^{40}$ and 0.83\% against AutoAttack.

\section{Experiments: Asymmetric DH-AT with ResNet}

We further evaluate our DH-AT strategy on the CIFAR-100 dataset using its asymmetric variant, \textit{i.e.}, the main head is a ResNet-34 and the second head is a revised Resnet-18 (since it has one more residual block than its standard version). We select SAT and TRADES as the baseline adversarial training methods and report the best checkpoint's performance. For robustness evaluation, we use the PGD$^{20}$ (no random restarts), PGD$^{100}$ (with 5 random restarts), and AutoAttack (AA). All attacks are bounded by $\epsilon=8/255$ and the step size of PGD (including its variants in AA) is set to $1/255$.

\begin{table}[h]
\caption{Robustness results on CIFAR-100 for different defense methods under various hyperparameter settings. All baseline methods (Natural, SAT and TRADES) use the ResNet-34 model. All DH-AT models utilize ResNet-34 as the main head and a revised ResNet-18 as the second head.}
\centering
\setlength{\tabcolsep}{0.56em}
\begin{tabular}{lcccc}
\toprule
                             & \textbf{Natural} & \textbf{PGD$^{20}$} & \textbf{PGD$^{100}$} & \textbf{AA} \\ \midrule
Natural Training            & \textbf{78.56\%}                       & 0.02\%                      & 0.00\%                          & 0.00\%                           \\ \midrule
Natural (DH w/ TRADES) & 73.86\%      & \textbf{14.89\%}    & \textbf{13.63\%}     & \textbf{10.65\%}       \\ \midrule
Natural (DH w/ SAT)   & 74.64\%      & 13.27\%    & 12.39\%     & 9.71\%      \\ \midrule \midrule

TRADES~\cite{zhang2019theoretically}                         & 58.13\%                       & 27.59\%                     & 26.37\%                      & 26.18\%                       \\ \midrule
TRADES (DH)       & 59.57\%      & \textbf{28.91\%}    & \textbf{27.28\%}     & \textbf{26.36\%}      \\ \midrule
TRADES (DH w/ PGD$^{2}$) & \textbf{68.72\%}      & 22.23\%    & 21.31\%     & 20.30\%  \\ \midrule \midrule

SAT~\cite{madry2017towards}                        & 60.48\%                       & 24.48\%                     & 23.45\%                      & 23.31\%                       \\ \midrule
SAT (DH)           & 61.39\%      & \textbf{25.67\%}    & \textbf{24.41\%}     & \textbf{23.32\%}      \\ \midrule
SAT (DH w/ PGD$^{2}$)    & \textbf{71.25\%}      & 19.86\%    & 18.57\%     & 16.09\%      \\ \bottomrule
\end{tabular}
\label{tbl:cifar100_results}
\end{table}

We train our DH-AT models using adversarial training (SAT or TRADES) with three different strengths for the main head: 1) natural (clean) training, PGD$^{2}$ and PGD$^{10}$. 
For PGD$^{10}$ adversarial training, we follow the same setting as SAT or TRADES, while for PGD$^{2}$ adversarial training, the step size is set to $2/255$ with $\epsilon=4/255$. Similar to previous experiments, we select the best checkpoint for the main head. For natural training, we train the ResNet-34 model for 40 epochs with the CE loss. Then, we add a second head and a lightweight CNN to this naturally-trained network, and train both of them using PGD$^{10}$ adversarial training.
The use of naturally-trained models as the main head is to demonstrate the effectiveness and practicability of our DH-AT strategy in more complex real-world defense scenarios. For example, the naturally-trained main head can be easily extracted from our DH-AT model to achieve high clean accuracy, while the adversarially-trained second head can be switched on to obtain robustness.

The results are reported in Table \ref{tbl:cifar100_results}.
Compared with the naturally-trained model, using DH-AT with TRADES for the second head can improve the robustness from 0.02\% to 14.89\% against PGD$^{20}$ attack, while the clean accuracy drops by only 4.7\% to 73.86\%. Additionally, higher robustness can be achieved by applying DH-AT with a stronger adversary for the main head. When using DH-AT with SAT where the main head is trained with a PGD$^{2}$ adversary, the clean accuracy can be improved by 10.77\%, while the robustness drops by 4.62\%. Ultimately, utilizing DH-AT on the standard TRADES can improve the robustness by 1.32\%, and at the same time, improve the clean accuracy by 1.44\%.
While there is still large room for improvement, these results demonstrate the effectiveness and flexibility of our DH-AT strategy to meet the diverse accuracy and robustness requirements of real-world applications.

\section{Conclusion}

In this work, we proposed a Dual Head Adversarial Training (DH-AT) strategy to combine different levels of adversarial robustness into one single model. DH-AT introduces both architectural and training modifications to existing deep neural networks (DNNs) and adversarial training methods.
The two heads in DH-AT models can be trained differently to improve the overall robustness, while maintaining or slightly improving clean accuracy. We showed that our DH-AT strategy can be readily implemented into different DNNs and adversarial training methods with minimal modifications.
Our proposed DH-AT strategy can be used as a practical tool to obtain both clean accuracy and adversarial robustness, or different levels of accuracy-robustness trade-off in a single model. 

\bibliographystyle{ieeetr}
\bibliography{ref}

\end{document}